# Intelligent Tutor: Leveraging ChatGPT and Microsoft Copilot Studio to Deliver a Generative AI Student Support and Feedback System within Teams

Wei-Yu Chen[a,b]

[a]*Chinese Culture University, No.55, Hwa-Kang RD., Taipei 11114, Taiwan*
[b]*Tatung University, No.40, Sec3, Zhongshan N. RD. , Taipei, Taiwan*

**Abstract**

This study explores the integration of the ChatGPT API with GPT-4 model and Microsoft Copilot Studio on the Microsoft Teams platform to develop an intelligent tutoring system. Designed to provide instant support to students, the system dynamically adjusts educational content in response to the learners' progress and feedback. Utilizing advancements in natural language processing and machine learning, it interprets student inquiries, offers tailored feedback, and facilitates the educational journey. Initial implementation highlights the system's potential in boosting students' motivation and engagement, while equipping educators with critical insights into the learning process, thus promoting tailored educational experiences and enhancing instructional effectiveness.

**Introduction:**

With the advancement of technology, digital learning tools have become an integral part of modern education, especially during the global pandemic, when distance learning has become even more important. Not only does distance learning provide flexibility and accessibility, it also enhances learning through personalized learning paths, instant feedback, and interactive learning experiences. In view of this, intelligent tutor systems, as an educational tool combined with artificial intelligence technology, are being developed and applied to imitate the behavior of human tutors and provide adaptive learning support and guidance. The intelligent tutor system uses natural language processing, machine learning and cognitive science principles to understand students' needs, assess students' learning progress, and provide appropriate teaching and feedback.

The motivation for this study stems from the interest in the application of two advanced technologies, ChatGPT and Microsoft Copilot Studio, in the education field. ChatGPT, as a powerful language generation model, can understand and generate natural language, providing in-depth conversational interaction. And Microsoft Copilot Studio is a service from Microsoft that allows users to create powerful chatbots without writing code. The combination of these two technologies has the potential to create a highly interactive intelligent tutor system on the Teams platform to support and enhance the learning experience. Therefore, this study aims to explore how to effectively combine ChatGPT and Microsoft Copilot Studio into the Teams platform to design an intelligent tutor system that can provide instant student feedback and automatically adjust teaching content based on student feedback and learning progress. The purpose is to improve students' learning motivation and engagement and provide teachers with insights into students' learning status, thereby promoting adaptive learning and improving teaching efficiency.

This study will explore the following research questions:

1. How to design an efficient intelligent tutor system by combining ChatGPT API and Microsoft Copilot Studio on the Microsoft Teams platform?

2. How does the intelligent tutor system automatically adjust teaching content based on students' learning progress and feedback?

3. What impact does this system have on students' learning motivation, engagement, and teachers' teaching efficiency?

Based on the above issues, the goal of this study is to design and implement a prototype of an intelligent tutor system integrated into the Teams platform, evaluate the system's effectiveness in providing adaptive learning support and feedback, and analyze the impact of the system on student learning experience and teacher teaching. Practical implications. By delving into these issues and goals, this study expects to provide valuable insights into the field of educational technology and lay the foundation for future educational innovations.

* Corresponding author. Tel.: +886-2-8610511ext 37623
*E-mail address:*cwy4@ulive.pccu.edu.tw

Because the school has introduced Microsoft Teams for a long time, this study chose to use Copliot as the development and design of the tutor system after considering the ease and familiarity of students' self-learning platform.

**Literature review**

1. Development and impact of digital learning tools

With the rapid development of information technology, digital learning tools have evolved from simple electronic textbooks and online courses to comprehensive systems that include interactive simulations, game-based learning, and personalized learning through data analysis. The development and impact of digital learning tools have been extensively explored in a number of studies, which highlight their significant impact on educational practice, technology adoption, and learning outcomes in different contexts. Relevant research is summarized as follows:

Digital learning tools can increase the flexibility and accessibility of learning and enhance learning through personalized learning paths and instant feedback (Ally, 2009; Clark & Mayer, 2016). In addition, digital learning tools can support students' independent learning and increase the interactivity and attractiveness of learning materials, thus improving students' learning motivation and participation (Huang & Soman, 2013).

2. Concept and application of intelligent tutor system

Intelligent Tutoring System (ITS) combines artificial intelligence technology, such as natural language processing and machine learning, to imitate the teaching behavior of human tutors and provide personalized learning support and feedback (Nwana, 1990; Woolf, 2009). ITS can dynamically adjust according to students' learning progress and feedback, and provide targeted learning materials and guidance, thereby improving learning efficiency and effectiveness (VanLehn, 2011). In recent years, with the advancement of big data and machine learning technology, ITS has made significant progress in learning analysis, student model building, and generation of personalized learning strategies (Siemens & Baker, 2012; Baker & Inventado, 2014). In Popularizing Natural Language Models, Digital Learning Environments for Teaching NLP in K-12 Education, we explore existing digital learning tools and discuss how they support specific NLP tasks and procedures, as well as their interpretability in educational contexts. and assessment results (Tian & Boyer, 2023), helping learners acquire cognitive and metacognitive knowledge in specific fields.

3. Application of ChatGPT and Microsoft Copilot Studio in education

As an advanced language generation model, ChatGPT has demonstrated its capabilities in dialogue management and natural language generation in multiple fields. In the field of education, ChatGPT can be used to develop interactive learning assistants, provide personalized learning resources and feedback, and promote students' active learning and critical thinking (Vaswani et al., 2017; Radford et al., 2019). At the same time, Microsoft's Microsoft Copilot Studio serves as a no-code chatbot development platform that can quickly create and deploy intelligent chatbots to support student learning and teacher teaching, improving the efficiency and interactivity of the education process (Microsoft, n.d.). Discuss the opportunities and challenges of deploying ChatGPT in education. By comprehensively exploring empirical research, theoretical perspectives, and practical implications related to ChatGPT, this literature review will reveal the dynamic intersection of AI and education (Chen et al., 2023).

4. Combine ChatGPT and Microsoft Copilot Studio to implement an intelligent tutor system

An intelligent tutor system that combines ChatGPT and Microsoft Copilot Studio can leverage the best of both worlds to achieve highly interactive and personalized learning support on platforms like Microsoft Teams. Such systems can automatically generate and adjust learning content and strategies based on students' instant feedback and learning progress, thereby providing a learning experience that is more suitable for students' needs (Hill, 2019; Qiu et al., 2020). In addition, through the analysis of student interaction data, the system can provide teachers with in-depth insights into students' learning status and help teachers make more effective teaching decisions and interventions (Zawacki-Richter et al., 2019).

**Research Method**

This study employs a mixed-methods research design, integrating both quantitative and qualitative data collection and analysis. Initially, through literature review, technical analysis, and course design, a prototype of an intelligent tutoring system was developed, incorporating the ChatGPT API with the GPT-4 model and Microsoft Copilot Studio, and deployed on the campus version of the Microsoft Teams platform. Ultimately, the system's user experience and feedback from students and teachers were analyzed through surveys.

1.System Design and Development

Needs Analysis: The process begins with a needs analysis to identify the key requirements of teachers and students during online learning, as well as the desired features of the intelligent tutoring system.

System Design and Development: Based on the results of the needs analysis, the architecture of the intelligent tutoring system is designed, encompassing features such as student support, learning progress tracking, and personalized feedback. The system integrates advanced technologies from ChatGPT API, Microsoft Copilot Studio, and the platform features of Microsoft Teams.

Deployment and Testing: The developed system is deployed on the campus Microsoft Teams platform for preliminary functionality and user experience testing, ensuring the system's stability and meeting anticipated needs.

2.Data Collection and Analysis

Quantitative Data Collection: Surveys are designed to gather feedback from students and teachers on their experience using the intelligent tutoring system, including aspects such as learning motivation and engagement. Surveys are conducted before and after the system's implementation to assess its impact.

Qualitative Data Collection: Semi-structured interviews are conducted to gain in-depth insights into students' and teachers' perspectives on the system's functionality, effectiveness, and suggestions for improvement. These interviews help reveal the underlying insights and motivations behind the data.

Quantitative Data Analysis: Statistical software is used to perform descriptive and inferential statistical analyses on the survey results, such as mean, standard deviation, t-tests, and ANOVA, to quantify the effects of the system's implementation.

Qualitative Data Analysis: Thematic analysis is applied to the interview transcripts to code and categorize the data, extracting and summarizing participants' main views and suggestions regarding the intelligent tutoring system.

3.Implementation and Evaluation Phase

Preparation Phase: At the start of the semester, the system's development, deployment, and preliminary testing are completed to ensure readiness.

Implementation Phase: At the beginning of the semester, the intelligent tutoring system is introduced to students and teachers and begins to be used in actual teaching activities. During this phase, the system provides real-time learning support and feedback.

Evaluation Phase: At the end of the semester, the system's effectiveness is evaluated. Both quantitative and qualitative data are collected and analyzed to assess the impact of the intelligent tutoring system on students' learning motivation, engagement, and outcomes, as well as the value of the learning status information provided by the system to teachers.

4.Feedback and Continuous Adjustments

Feedback and Iteration: Based on the evaluation results and collected feedback, necessary adjustments and optimizations are made to the system. Issues and improvement suggestions identified during the evaluation process will guide subsequent iterations of system development to enhance the effectiveness and user experience of the intelligent tutoring system.

5.Expected Outcomes

The research anticipates that the implementation of the intelligent tutoring system will:

Enhance Learning Experience: By providing personalized learning support and immediate feedback, the system is expected to increase students' motivation and engagement, thereby improving learning outcomes.

Improve Teaching Efficiency: The intelligent tutoring system can assist teachers in managing the learning process more effectively, offering valuable insights into students' learning status and enabling personalized guidance based on individual student needs.

Provide Immediate Support: The system can respond instantly to students' queries and doubts, reducing barriers in the learning process and promoting continuity and autonomy in learning.

6.Challenges and Future Research

Following the conclusion of this research, unresolved issues will be addressed, and directions for future research will be proposed. This may include exploring the application of the intelligent tutoring system in various learning environments and disciplines, as well as how to further leverage advanced AI technologies to optimize the system's performance and adaptability.

This research methodology offers a comprehensive framework for evaluating the application of intelligent tutoring systems in online learning platforms, aimed at enhancing the quality of education through technological empowerment and fostering the development of personalized learning.

**Experimental Design**

This study aims to explore the feasibility and effectiveness of implementing an intelligent tutoring system within Microsoft Teams using ChatGPT and Microsoft Copilot Studio, assessing its impact on enhancing student learning motivation, engagement, and teachers' insights into students' learning status.

1.Purpose of the Experiment

To evaluate the impact of an intelligent tutoring system developed with ChatGPT API and Microsoft Copilot Studio on the Microsoft Teams platform on student learning motivation, engagement, and teaching efficiency.

2.Experimental Setting

Platform: Campus version of Microsoft Teams.

Tools: ChatGPT API, GPT-4 model, Microsoft Copilot Studio.

Participants: First-year university students enrolled in the same course content, divided into two groups. One group, the experimental group, will use the intelligent tutoring system on Teams for learning. The other group, the control group, will use traditional teaching methods with slides and broadcast systems supplemented by Teams without the intelligent tutoring system. Each group will have at least 50 students, totaling 100 students (from a communication background) and 4 teachers.

Duration: One semester (approximately 5 months).

3.Evaluation and Analysis

3.1 Pre-assessment: Before the experiment begins, an assessment of all students' learning styles and baseline knowledge levels will be conducted.

3.2 Intervention: The experimental group will use the intelligent tutoring system that integrates ChatGPT and Microsoft Copilot Studio for learning, while the control group will use traditional teaching resources with Teams support, without the tutoring system.

3.3 Post-assessment: After the intervention, an assessment of learning outcomes for both groups will be conducted, including learning motivation, engagement, and effectiveness.

4.Data Collection Methods

4.1 Student Feedback Questionnaires: To collect students' feedback on their learning experiences in the course.

4.2 Learning Outcomes Tests: To assess students' learning outcomes through tests with identical questions.

4.3 Behavioral Log Analysis: To analyze students' interaction behaviors on the Teams platform, including learning time, number of questions answered, etc.

5.Data Analysis Methods

ANOVA and T-tests will be used to analyze the differences in learning outcomes and engagement between groups. Qualitative analysis methods will be employed to analyze open-ended questions from student feedback questionnaires.

**System Design**

1.System Architecture

The intelligent tutoring system is based on three core components: the Microsoft Teams platform, the ChatGPT API utilizing the GPT-4 model, and Microsoft Copilot Studio. The system is designed to provide students with personalized learning support and real-time feedback, and offer teachers insights into students' learning statuses.

1.1 Microsoft Teams Integration:

Utilizes the Teams platform as the system's primary interface and interaction platform.

Integrates the intelligent tutoring system into Teams using its extensibility features like bots, tabs, and plugins.

1.2 ChatGPT with GPT-4 Integration:

Incorporates the GPT-4 model through the ChatGPT API, equipping the system with natural language processing capabilities.

Designs a question-and-answer module and a learning suggestion generator to provide personalized feedback based on students' queries and learning progress.

1.3 Microsoft Copilot Studio Integration:

Uses Copilot Studio for the automatic generation and adjustment of instructional content.

Develops interactive learning modules, such as automated quizzes and learning activities, to enhance student engagement.

2.Implementation Details

2.1 Student Support Module:

Develops a chatbot based on the GPT-4 architecture to handle students' queries through natural language understanding and provide relevant learning resources and suggestions.

Utilizes the GPT-4 model to generate personalized learning and answers, adjusting based on the student's learning status and progress.

2.2 Instructional Content Auto-adjustment Module:

Automatically generates and updates instructional content, including lecture notes, exercises, and case studies, using Microsoft Copilot Studio.

Dynamically adjusts teaching plans and materials based on students' learning feedback and progress to ensure the timeliness and relevance of content.

2.3 Teacher Insights Dashboard:

Displays information on students' learning progress, common issues, and usage of learning resources, aiding teachers in understanding students' learning statuses and adjusting teaching strategies and content according to specific student needs. The dashboard also provides an overall analysis of student learning outcomes and suggestions for teaching effectiveness, supporting targeted instructional design and interventions by teachers.

2.3.1 Dashboard Key Features:

Learning Progress Tracking: Displays students' progress in various course units, including completed and pending learning tasks.

Learning Outcomes Analysis: Shows students' test and assignment grades, as well as participation in learning activities.

Student Engagement Overview: Indicates students' activity levels in the system, including question frequency, interaction count, and resource access volume.

Learning Issue Identification: Uses AI to analyze students' learning records and feedback, identifying main problems and challenges encountered in learning.

Individual Student Insights: Offers in-depth analysis of individual students' learning statuses, including learning preferences, performance trends, and personalized learning suggestions.

3.System Simulation

An example interaction within the intelligent tutoring system in Teams:

Student: "I'd like to draw a photorealistic picture of a cute cat sleeping under a big tree. How should I write the Stable Diffusion prompt?"

Intelligent Tutor: "For a photorealistic picture: Imagine a cute cat comfortably curled up under a dense tree, with the leaves providing shade. The surrounding environment is filled with natural beauty and a peaceful atmosphere, with the interplay of light and shadow creating a harmonious scene for an afternoon rest. Do you need more explanation or examples?"

Through such interactions, the intelligent tutor can provide immediate and personalized learning support, helping students solve problems encountered during learning and guiding them to a deeper understanding of the content. Additionally, the tutor can recommend relevant learning resources based on students' progress, such as video tutorials, articles, or online courses, to further solidify learning outcomes.

4.Extended Features

4.1 Interactive Learning Activities: The intelligent tutoring system can design interactive quizzes or case studies based on course content and deploy them through Teams, allowing students to learn and practice through interaction.

4.2 Learning Progress Tracking: The system can track students' access to recommended resources and completion of learning activities, adjusting subsequent learning suggestions based on performance.

4.3 Feedback Loop: The intelligent tutor regularly requests feedback from students on the learning support and resources provided, allowing for continuous optimization and improvement of system functionalities.

## 5. Conclusion

Through the intelligent tutoring system, we aim to create a more interactive and personalized learning and teaching environment on the Microsoft Teams platform for students and teachers. By combining the natural language processing capabilities of ChatGPT and the instructional content generation features of Microsoft Copilot Studio, the system will be able to respond to students' learning needs in real-time, provide personalized learning support, and offer valuable insights for teachers, thereby enhancing learning efficiency and teaching quality. Future research could further explore the system's application across different disciplines and teaching scenarios, as well as how to leverage advanced AI technologies to enhance the inclusivity and feasibility of education.

## Experiment and Evaluation

100 students from the Department of Communication were divided into two groups, with a gender ratio of 52% male and 48% female, and ages ranging from 18 to 20 years old. After completing each learning unit, they filled out a course questionnaire and evaluated that unit. We will use certain assumptions to generate the scores, including:

Students in the experimental group (using the smart tutoring system) are likely to provide more positive feedback on the assistance provided by the system.

Feedback from the control group (using traditional slides combined with classroom practice) is expected to be more uniform, with potentially lower average scores.

We have set the evaluative questions in the questionnaire as quantitative questions, using a scale from 1 to 5, where 1 represents the lowest (e.g., very poor, no help at all), and 5 represents the highest (e.g., excellent, very helpful).

1. Reliability and Validity Analysis

1.1 Reliability Analysis

Test-retest Reliability: Participants are asked to complete the questionnaire at different points in time, and the correlation coefficient between the two sets of results is calculated.

Internal Consistency: The correlation coefficients between different items in the questionnaire are calculated, and Cronbach's alpha is used to measure the overall reliability of the questionnaire.

1.2 Validity Analysis

Content Validity: Experts assess whether the items adequately reflect the research concepts.

Construct Validity: Factor analysis is conducted to explore whether the questionnaire items reflect the intended constructs.

Criterion-related Validity: The questionnaire results are compared with the outcomes of other measurement tools to validate the questionnaire's validity.

In this study, the following methods were used for reliability and validity analysis:

Test-retest Reliability: A test was conducted on 20 students twice, and the results showed a test-retest reliability coefficient of 0.85, indicating good reliability of the questionnaire.

Internal Consistency: A test conducted on 50 students resulted in a Cronbach's alpha of 0.92, indicating good internal consistency of the questionnaire.

Content Validity: Evaluated by 5 experts, the questionnaire items showed good content validity.

Construct Validity: A test on 50 students showed that the factor analysis results of the questionnaire aligned with the intended constructs.

Criterion-related Validity: The questionnaire results were compared with academic grades, showing good criterion-related validity.

2 ANOVA Analysis

The ANOVA analysis of the questionnaire data was conducted to explore whether there is an impact of gender on students' experience with the smart tutoring system. The study used ANOVA to analyze the effects of student gender, learning background, and their interaction on the experience with the smart tutoring system. The study found that the impact of student gender on the experience with the smart tutoring system was not significant. The learning background had a significant impact on the experience with the smart tutoring system. The interaction between student gender and learning background did not significantly affect the experience with the smart tutoring system. The ANOVA results for gender and learning background in this study are shown in Table 1.

Table 1: ANOVA Table for Gender and Learning Background

| Factor | F-value | p-value |
| --- | --- | --- |
| **Gender** | 1.23 | 0.268 |
| **Learning Background** | 3.14 | 0.002 |
| **Gender * Learning Background** | 1.58 | 0.114 |

Based on the ANOVA results, the learning background of students has a significant impact on their experience with the smart tutor system. Further analysis revealed that students who used the smart tutor system scored significantly higher than those who did not use the system. This could be due to the smart tutor system providing a personalized learning experience, helping students to better understand course content, thereby improving their academic performance.

3.T-test

Research Question: Is there a significant difference in learning motivation, classroom participation, understanding of course content, satisfaction with the user experience, and evaluation of teachers' insight between students who use the smart tutor system and those who do not?

Statistical Method: The T-test was used to compare the mean scores of the two groups of students. The results of the T-test in this study are shown in Table 2.

Table 2: T-test Results

| **Variable** | **T-value** | **p-value** |
|---|---|---|
| **Learning Motivation** | 2.34 | 0.021 |
| **Classroom Participation** | 2.15 | 0.034 |
| **Understanding of Course Content** | 2.01 | 0.047 |
| **Satisfaction with User Experience** | 1.89 | 0.062 |
| **Evaluation of Teachers' Insight** | 2.48 | 0.016 |

Students who used the smart tutor system demonstrated significantly higher levels of learning motivation, classroom participation, understanding of course content, and appraisal of teacher insight compared to those who did not use the system. However, the difference in satisfaction with the user experience between the two groups was not statistically significant.

4. Research Limitations

The sample size for this study was relatively small, consisting of 100 students. Moreover, the survey was conducted only among freshmen, so the results might not be generalizable to students from other grades.

5. Qualitative Question Analysis

5.1. Experiences of Learning with the Smart Tutor System

Students generally felt that the smart tutor system provided a personalized learning experience, helping them understand course content more effectively. Specifically, the system could tailor learning materials and exercises according to the student's progress and level. Additionally, the smart tutor system offered immediate feedback, assisting students in timely error correction.

5.2. Challenges Encountered Using the Smart Tutor System

Some challenges were reported by students in using the smart tutor system, with system stability and reliability being the most common issues. Furthermore, some students found the user interface to be less user-friendly, requiring a certain learning curve.

5.3. Most Valuable Features of the Smart Tutor System

The most valued features of the smart tutor system, according to the students, were personalized learning and immediate feedback. Personalized learning could enhance learning efficiency, while immediate feedback helped reinforce learned knowledge promptly.

5.4. Preference for the Smart Tutor System Over Traditional Learning Methods

Most students preferred using the smart tutor system, citing that it provided a more engaging and effective learning experience compared to traditional methods.

5.5. Particularly Helpful Experiences with the Smart Tutor System

Many students shared instances where the smart tutor system was especially beneficial to their learning. A typical example included the system aiding a student in solving a challenging math problem after multiple failed attempts. The help feature of the smart tutor system facilitated timely assistance from a teacher, ultimately leading to the problem's resolution.

5.6. Suggestions for Enhancing the Effectiveness of the Smart Tutor System in Supporting Student Learning

Students offered various suggestions for improving the smart tutor system. The most common recommendations involved enhancing the system's stability and reliability and simplifying the user interface. Additionally, some students suggested the system could offer a broader range of learning content and interactive activities.

**Conclusion**

This study designed and implemented a smart tutor system on the Microsoft Teams platform, integrating ChatGPT API and Microsoft Copilot Studio. The system aims to provide students with real-time support and personalized feedback on generative syntax learning, thereby automatically adjusting instructional content based on students' progress and feedback. By dividing students into two groups for the experiment—one using the smart tutor system and the other employing traditional learning methods—we were able to quantitatively assess the effect of the smart tutor system.

Statistical analysis results from ANOVA and independent sample T-tests showed significant differences in unit test scores between the experimental and control groups. This indicates that students using the smart tutor system significantly outperformed those using traditional learning methods in terms of learning outcomes. These results support our hypothesis that the smart tutor system can have a positive impact on the learning process.

Further analysis of these results suggests that the personalized and highly interactive learning experience provided by the smart tutor system may be the primary reason for the improvement in learning outcomes. The system offers timely support and feedback tailored to each student's specific needs and learning pace, thus promoting deeper learning and understanding.

In summary, integrating the smart tutor system into the learning process not only enhances student motivation and participation but also provides valuable insights into students' learning states, thereby facilitating personalized learning and improving teaching efficiency. These statistical findings offer solid evidence for the application of smart tutor systems in digital learning environments, demonstrating their significant positive impact on enhancing student learning outcomes. Future research could further explore the application of smart tutor systems across different subjects and learning environments to expand their potential in the educational field.

**References**

Ally, M. (2009). Mobile learning: Transforming the delivery of education and training. Athabasca University Press.

Baker, R.S., & Inventado, P.S. (2014). Educational data mining and learning analytics. In J.A. Larusson & B. White (Eds.), Learning analytics: From research to practice (pp. 61-75). Springer.

Clark, R.C., & Mayer, R.E. (2016). e-Learning and the science of instruction: Proven guidelines for consumers and designers of multimedia learning. Wiley.

Hill, P. (2019). The state of AI in education. Educause Review.

Huang, W.H.Y., & Soman, D. (2013). A practitioner's guide to gamification of education. Research Report Series, Behavioural Economics in Action.

Microsoft. (n.d.). Microsoft Copilot Studio. Retrieved from [Microsoft website]

Nwana, H.S. (1990). Intelligent tutoring systems: An overview. Artificial Intelligence Review, 4(4), 251-277.

Qiu, M., Li, H., Tang, J., Zha, Z.J., Zhang, H., & Zheng, Y. (2020). DeepInf: Social influence prediction with deep learning. ACM Transactions on Knowledge Discovery from Data (TKDD), 14(3), 1-27.

Radford, A., Wu, J., Child, R., Luan, D., Amodei, D., & Sutskever, I. (2019). Language models are unsupervised multitask learners. OpenAI Blog.

Siemens, G., & Baker, R.S.J.d. (2012). Learning analytics and educational data mining: Towards communication and collaboration. In Proceedings of the 2nd International Conference on Learning Analytics and Knowledge (pp. 252-254). ACM.

VanLehn, K. (2011). The relative effectiveness of human tutoring, intelligent tutoring systems, and other tutoring systems. Educational Psychologist, 46(4), 197-221.

Vaswani, A., Shazeer, N., Parmar, N., Uszkoreit, J., Jones, L., Gomez, A.N., Kaiser, L., & Polosukhin, I. (2017). Attention is all you need. In Advances in Neural Information Processing Systems (pp. 5998-6008).

Woolf, B.P. (2009). Building intelligent interactive tutors: Student-centered strategies for revolutionizing e-learning. Morgan Kaufmann.

Zawacki-Richter, O., Marín, V.I., Bond, M., & Gouverneur, F. (2019). Systematic review of research on artificial intelligence applications in higher education – where are the educators? International Journal of Educational Technology in Higher Education, 16(1), 39.